\pdfoutput=1
\documentclass[journal]{IEEEtran}
\usepackage{hyperref}
\usepackage{framed,multirow}
\usepackage{cite}
\usepackage{amsmath, amssymb}
\usepackage{amsthm}
\usepackage{graphicx}
\usepackage{amsfonts}
\usepackage{longtable}
\usepackage{epsf, epsfig}
\usepackage{epstopdf}
\usepackage{subfigure}
\usepackage{algorithm}
\usepackage{algorithmic}
\usepackage{color}
\usepackage[normalem]{ulem}
\usepackage{booktabs}
\usepackage{adjustbox}
\usepackage{xspace}
\usepackage{doi}

\usepackage{xargs} 
\usepackage[pdftex,dvipsnames]{xcolor}



\graphicspath{{images/}}

\newcommand{\eat}[1]{}

\newcounter{gaocomm} 
  
\definecolor{blue-violet}{rgb}{0.54, 0.17, 0.89}
\definecolor{mygreen}{rgb}{0.0, 0.5, 0.0}
\definecolor{awesome}{rgb}{1.0, 0.13, 0.32}
\definecolor{bostonuniversityred}{rgb}{0.8, 0.0, 0.0}

\begin{document}

\title{Enhancing Generalizability of Predictive Models with Synergy of Data and Physics}

\author{
\thanks{Corresponding author: Zhe Song.}
\thanks{Y. Shen is with the Business School, Nanjing University, Nanjing, 210093, China (e-mail: dg1902054@smail.nju.edu.cn).}
\thanks{Z. Song is with the Business School, Nanjing University, Nanjing, 210093, China. (e-mail: zsong1@ nju.edu.cn).}
\thanks{A. Kusiak is with the Department of Industrial and Systems Engineering, 4627 Seamans Center for the Engineering Arts and Sciences, The University of Iowa, Iowa City, IA, USA (e-mail: andrew-kusiak@uiowa.edu).}

\normalsize{
Yingjun Shen,
Zhe Song\textsuperscript{$\ast$},~\IEEEmembership{Member, IEEE}, 
Andrew Kusiak,~\IEEEmembership{Life Member, IEEE}
}
}


\maketitle
\begin{abstract}
    Wind farm needs prediction models for predictive maintenance. There is a need to predict values of non-observable parameters beyond ranges reflected in available data. A prediction model developed for one machine many not perform well in another similar machine. This is usually due to lack of generalizability of data-driven models. To increase generalizability of predictive models, this research integrates the data mining with first-principle knowledge. Physics-based principles are combined with machine learning algorithms through feature engineering, strong rules and divide-and-conquer. The proposed synergy concept is illustrated with the wind turbine blade icing prediction and achieves significant prediction accuracy across different turbines. The proposed process is widely accepted by wind energy predictive maintenance practitioners because of its simplicity and efficiency. Furthermore, this paper demonstrates the importance of embedding physical principles within the machine learning process, and also highlight an important point that the need for more complex machine learning algorithms in industrial big data mining is often much less than it is in other applications, making it essential to incorporate physics and follow “Less is More” philosophy.
\end{abstract}


\section{Introduction}\label{Sec:1}
Manufacturing becomes more and more intelligent with the development of Big Data, Cloud Computing, Internet of Things, and Artificial Intelligence ~\cite{kusiak2017smart}.  Sensors are installed to collect data at all stages of production, such as material properties, temperatures and vibrations of equipments and customer characteristics~\cite{jeschke2017industrial,da2014internet}. The digital manufacturing transformation has led to more sensors generating large volumes of data which can be mined for many purposes. For example, engineers can improve product quality, optimize existing processes, predict emerging failures based on historical data ~\cite{helbing2018deep,kusiak2015big,kusiak2015break,mcafee2012big,stetco2019machine}.

How to build generalizable predictive models is one of the major challenges in machine learning research. Predictive models should be able to handle uncertainties and generalize before industrial deployment~\cite{kusiak2017four,kusiak2017smart}. Prediction models in industrial fields are often characterized as highly nonlinear and high-dimensional.

Generally speaking, there are two ways of building prediction models for complex industrial systems~\cite{kuo2019data}. One is top-down approach. The other is bottom-up approach. Top-down approach usually refers to constructing models by existing knowledge, such as Newton’s laws, aerodynamics, chemical reactions and so on. Top-down approach is also known as white-box model, or mechanistic model which is explainable, understandable and generalizable. Mechanistic model usually can’t predict well in high-resolution because of “random noises” in reality. Sometimes engineers work with problems for which there is no simple or well-understood mechanistic model that explains the phenomenon. Pure physics-based model will fall short in practical industrial setting. For example, Betz law may tell you the exact relationship between wind speed and the energy captured by a wind turbine. But 100 hundred same-type wind turbines will show 100 hundred different power curves drawn from field data~\cite{kusiak2016renewables}.

Bottom-up approach is usually related with data-driven methods. Models are derived from data by data mining or machine learning algorithms, which are sometimes called black boxes or empirical models~\cite{tan2016introduction,tao2018data}. For example, artificial neural networks are used to predict gearbox failures. Deep learning algorithms are used to predict wind turbine blade breakage ~\cite{wang2016wind,wu2013data}. Not like mechanistic model, data-driven empirical model may not generalize, but can make accurate predictions locally (somehow memorize the data you feed into the learning algorithm). As you can imagine, a stable production system maintains the controllable process variables, such as temperatures, as closely as possible to desired targets or set points. Because the controllable variables change so little, it may be difficult to assess their real impact on target response variables. On the other hand, in industrial scenario, data mining may involve a lot of data, but that data may contain relatively little useful information about the whole picture of the problem. Data is usually unbalanced because system runs in several steady states. Furthermore, some of the relevant data may be missing, there may be transcription or recording errors resulting in outliers, or data on other important factors may not have been collected and archived.

In reality, it happens to predict beyond specific conditions where data may be available. For these reasons, a purely data-driven prediction model won’t work. 

Achieving generalizable predictive models for these complex systems requires a synergistic combination of data and physics-based knowledge~\cite{ali2018online,lu2015modeling}. Learning from data through the lens of physical principle is a way to bring generalizability to an otherwise intractable prediction modeling problem. It is a way to integrate physical laws and domain knowledge. So, one promising direction is hybrid modeling. But how to embed the knowledge into the modeling process is an open research question.

Although there are hundreds of different machine learning algorithms available to build predictive models, few algorithms are designed specifically for improving model generalizability. In other words, algorithms alone are hard-pressed to learn a generalizable predictive model. However, if machine learning is viewed as a production process and final products are predictive models, new research efforts could be focused on optimizing the learning process. By scrutinizing the process and fixing potential loopholes, it is possible to wringing every last drop of value from industrial big data~\cite{bengio2013representation,pan2009survey,settles2009active,zheng2015methodologies}.

\begin{figure}[t]
\begin{minipage}[t]{0.8\linewidth}
\centering
\includegraphics[width=\linewidth]{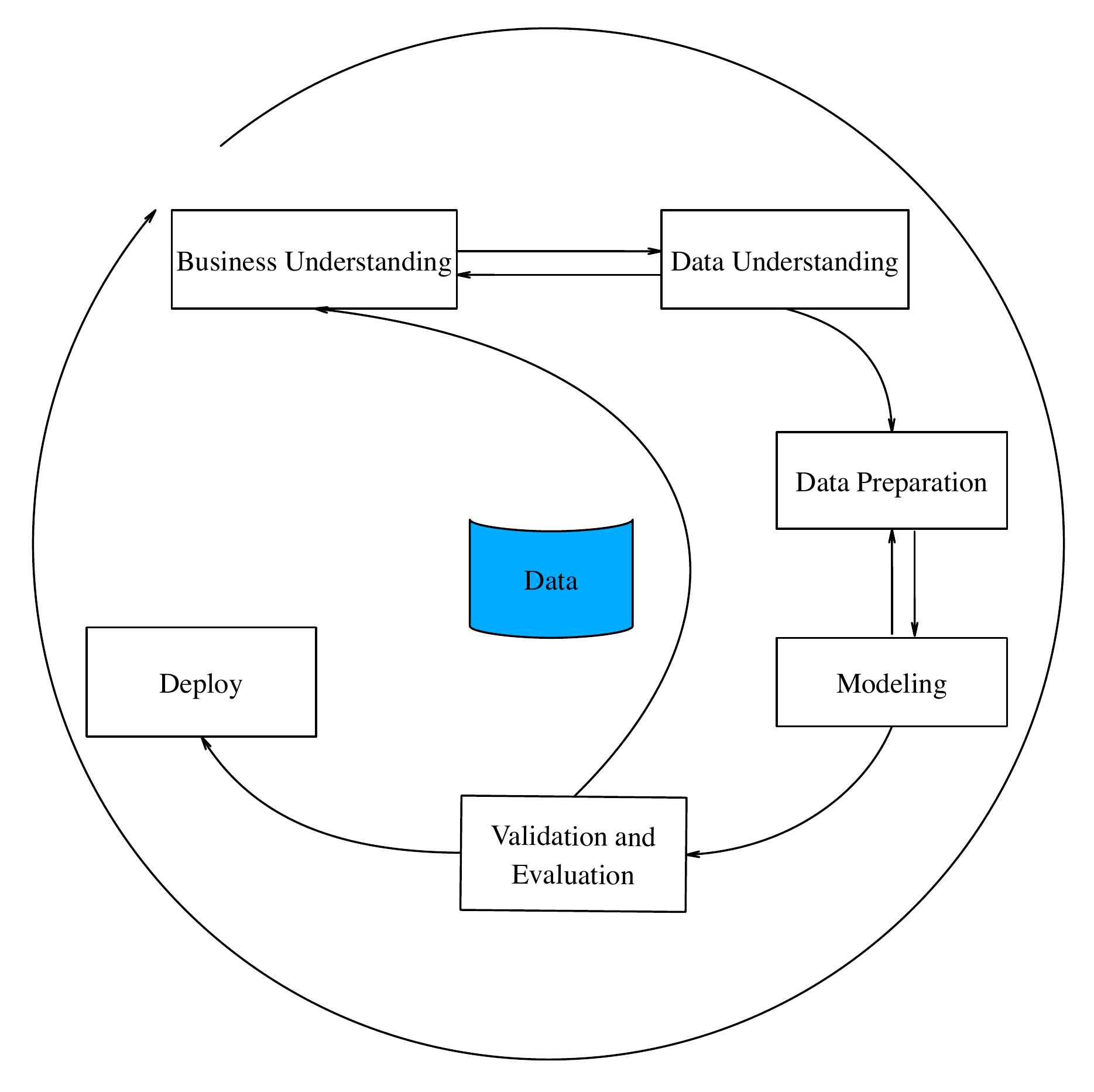}
\end{minipage}
\caption{CRISP-DM process}
\label{fig:CRISP-DM}
\end{figure}

Generally speaking, machine learning process can be divided into several parts: business understanding and data preparation, feature engineering, machine learning algorithm selection or parameter tuning, training, validation and testing, model deployment. Cross-Industry Standard Process for Data Mining (CRISP-DM) is a standard data mining process initiated in 1999, which is adopted by industry and academia for nearly 20 years with little innovation (Fig.~\ref{fig:CRISP-DM}) . Recent research about data mining and machine learning for predictive modeling is usually focused on algorithms innovation, such as deep learning algorithms~\cite{wu2013data}.

In existing literatures, predictive models are extracted from data by supervised learning algorithms. Generalizability is considered by choosing simpler models or increasing the data set size. Specific examples are early stopping or post-pruning of decision tree algorithms~\cite{sun2014predicting,tan2016introduction}. In this paper, generalizability of predictive models is improved through process renovation. Just like ImageNet solved the image recognition accuracy problem not by creating new machine learning algorithms, but by growing high-quality labeled image database. Physical principle is embedded into the learning process and new sub-processes are created to improve the predictive model’s generalizability. Our contribution is not creating new machine learning algorithms, but enriching the traditional data mining process. Sub-processes are developed to integrate physical principles, such as generating useful features according to physical equations, dividing machine learning problem into sub-problems. Good features based on physical principles will significantly improve prediction model’s generalizability. Physical principles could guide us to divide the prediction model into several simpler sub-models, which could be learnt easily. This paper used real-world wind turbine blade icing prediction to validate the proposed new process.

This article is structured as follows: case study background and data description are presented in Section~\ref{Sec:2}. Methods and processes are introduced in  Section~\ref{Sec:3}.  Section~\ref{Sec:4} and  Section~\ref{Sec:5} describes the computational processes, experiments and competition results. Implications and limitations of the study are discussed in  Section~\ref{Sec:6}.

\section{Case Study Background and Data Description}\label{Sec:2}

Case study of this paper is based on \emph{China Industrial Big Data Innovation Competition in 2017}, which is the first authoritative national competition in the field of industrial big data under the guidance of Ministry of Industry and Information Technology. The competition’s goal is to help companies solve realistic industrial big data problems, improve the level of smart manufacturing. The competition problem is to build blade icing prediction model for Wind Turbines (WT) from historical Supervisory Control and Data Acquisition (SCADA) data. The authors participated in the competition and achieved the second place out of nearly 1500 teams.  The competition is based on blind tests which has proven the effectiveness of our method.

Predictive maintenance and prognosis health management with SCADA data is the prevailing strategy adopted by wind farm operators~\cite{stetco2019machine,susto2014machine,tautz2016using}. Prediction model is the key technology for wind power industry~\cite{kusiak2016renewables,qiao2015survey}. Wind turbine’s tower height is increasing with the rated power output~\cite{blanco2009economics}. In China northern coastal or mountainous areas, a large number of wind turbines will touch the lower clouds in winter, which is very easy to freeze in the low-temperature and humid environment. Blade icing is a great threat to wind power generation and safe operation. Experience shows that blade icing will change blade shape and undermine aerodynamic characteristics, resulting in the lower efficiency and unstable production~\cite{chang2014comparative,chou2013failure,etemaddar2014wind}. Predicting blade icing is to compare a turbine’s SCADA power curve with the theoretic one. If the SCADA power curve is significantly lower than the theoretic one, alarm is triggered and wind turbine is shut down. However, this approach isn’t working well in reality. Alarm is usually triggered too late and large area icing is already formed on the blades. The blades will suffer high risk of breakage and may cause disastrous accidents. Thus, it is necessary to have an accurate blade icing prediction model so that the alarm can be triggered in the early stage of icing and de-icing system is turned on.

The competition organizer provided 6-month SCADA data of five wind turbines (named as WTs 8, 10, 14, 15, and 21). Each wind turbine has 28 variables such as wind speed, power and environmental temperature, which are listed in Table~\ref{table:data set variables}. Historical SCADA data of WTs 15 and 21 are used for training. Icing and no-icing time periods are provided so that “normal” and “abnormal” tags could be labeled for WTs 15 and 21. Based on the labeled data, supervised machine learning algorithms are applied to build prediction models.

\begin{table}[t]
\tiny
		\caption{SCADA data set variables (features)}
		\label{table:data set variables}
		\centering
		\begin{tabular} {cc} \toprule
				\textbf{Variables}	&	\textbf{Description} \\ \midrule
				time & Time stamp  \\   
				wind\_speed	 &   Wind speed	\\ 
				generator\_speed	 &   Generator speed	\\ 
				power	 &   Active power on network side	\\ 
				wind\_direction	 &   Opposite wind angle 	\\ 
				wind\_direction\_mean	 &   25 seconds average wind direction	\\ 
				yaw\_position	 &   Yaw position	\\ 
				yaw\_speed	 &   Yaw speed   \\
				pitch1\_angle	 &   Blade 1 angle	\\ 
				pitch2\_angle	 &   Blade 2 angle	\\ 
				pitch3\_angle	 &   Blade 3 angle 	\\ 
				pitch1\_speed	 &   Blade 1 speed	\\ 
				pitch2\_speed	 &   Blade 2 speed	\\ 
				pitch3\_speed	 &   Blade 3 speed	\\ 
				pitch1\_moto\_tmp	 &   Pitch motor 1 temperature	\\ 
				pitch2\_moto\_tmp	 &   Pitch motor 2 temperature	\\ 
				pitch3\_moto\_tmp	 &   Pitch motor 3 temperature	\\ 
				acc\_x	 &   X-direction acceleration	\\ 
				acc\_y	 &   Y-direction acceleration	\\ 
				environment\_tmp	 &   Ambient temperature	\\ 
				int\_tmp	& Cabin temperature	\\ 
				pitch1\_ng5\_tmp	 &   Charger 1 temperature	\\ 
				pitch2\_ng5\_tmp	 &   Charger 2 temperature	\\ 
				pitch3\_ng5\_tmp	 &   Charger 3 temperature	\\ 
				pitch1\_ng5\_DC	 &   Current of charger 1	\\ 
				pitch2\_ng5\_DC	 &   Current of charger 2	\\ 
				pitch3\_ng5\_DC	 &   Current of charger 3	\\ 
				group	 &   Data grouping identification	\\ 
				\bottomrule
		\end{tabular}
\end{table}

WTs 8, 10 and14 are used for blind testing, where the icing and no-icing information is not provided in the data sets. WT 8 is used for preliminary testing and contestants can submit their prediction results on a daily basis. The organizer will calculate and rank prediction scores based on the submitted predictions. Contestants can use this score to evaluate their algorithms’ performance and make necessary adjustments.

WTs 10 and 14 are used for final testing, which is released only two days before final submission. Contestants will have only two chances of submitting predictions in two days. The final prediction score is based on WTs 10 and 14’s prediction accuracy.

The total datasets are composed of 1,124,741 observations collected between November 2015 and January 2016, where 584,380 observations are labeled (WTs 15 and 21). Table~\ref{table:competition datasets} shows details of 5 wind turbine SCADA datasets.

\begin{table}[t]
\tiny
		\caption{Description of competition datasets}
		\label{table:competition datasets}
		\centering
		\begin{tabular} { p{1.5cm}p{1.5cm}p{2.0cm}p{2.0cm} } \toprule
				\textbf{Wind turbine}	 &   \textbf{Total observations}	 &   \textbf{Icing observations}	 &   \textbf{Sampling frequency (seconds)} \\ \midrule
				15	 &   393886	 &   23892	 &   	7,8,10	\\
				21	 &   190494	 &   10638		 &   7,8,10	\\
				8	 &   202328	 &   Not disclosed	 &   Not disclosed	\\
				10	 &   174301	 &   Not disclosed	 &   Not disclosed	\\
				14	 &   163732	 &   Not disclosed	 &   Not disclosed	\\
				\bottomrule
		\end{tabular}
\end{table}
.

\section{Data Mining Process Renovation with Physical Principle}\label{Sec:3}

\begin{figure}[t]
\begin{minipage}[t]{1\linewidth}
\centering
\includegraphics[width=\linewidth]{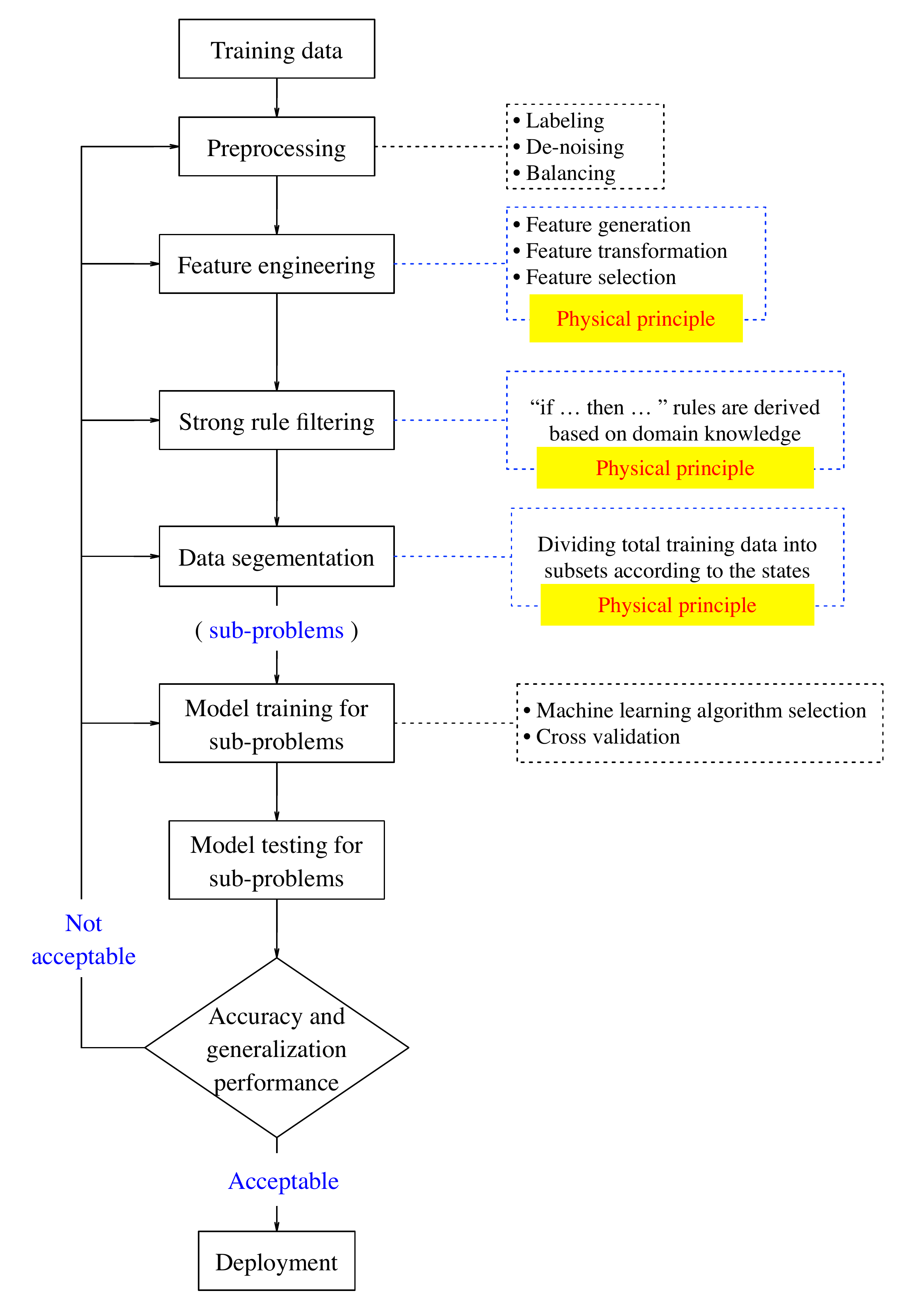}
\end{minipage}
\caption{Machine learning process for complex engineering problems with physical principle integration}
\label{ML_physical_principle}
\end{figure}

“Less is More” philosophy and Occam’s Razor theorem tell us to choose simple yet accurate model for engineering applications~\cite{guyon2003introduction,tan2016introduction}. In order to build generalizable prediction model, it is necessary to renovate existing data mining process by utilizing physical principle (see Fig.~\ref{ML_physical_principle}, three highlighted sub-processes). Previous research has shown that feature engineering is one of the most important machine learning process. Applied and real-world machine learning is basically feature engineering~\cite{sun2014predicting,zdravevski2017improving}. It is a perfect spot to embed physical principle into feature engineering.

For complex industrial big data mining, “Divide and Conquer” strategy is good at building more accurate and simpler prediction models. As most industrial systems are running in a set of different states, it is very reasonable to divide total training data into subsets according to these states. Then for each subset of training data, applying appropriate machine learning algorithm to learn the prediction model. As a result, it is natural to use different prediction models for different states respectively, which will have a better prediction performance than building only one prediction model. Following the “Divide and Conquer” strategy, it is necessary to build separate blade icing prediction models for low wind speed and high wind speed scenarios. It is noteworthy that “Divide and Conquer” strategy may fail if there are not enough training samples as you try to partition the training data set into too many categories.

Sometimes observations of training data set can be well explained or predicted by physical principles. In such circumstances, there is no need to use machine learning algorithm for prediction. Physical principle can be transformed into a set of “IF…THEN…” rules. For example, IF outside temperature is above some number, THEN it is impossible to have icing problem. IF wind speed is above some value AND power is above some value, THEN there is no blade icing. These rules could be called strong rules. Strong-rule filtering is a necessary process before letting machine learn. Machine learning algorithms are focused on data points which are not well explained by physical principle or domain knowledge.

Based on above discussions, traditional machine learning process is re-engineered to incorporate physical principle. Fig~\ref{ML_physical_principle} is the new machine learning process where feature engineering guided by physical principle, strong rule filtering and data segmentation (“Divide and Conquer”) are added. Like the CRISP-DM process, this reengineered process is not a pure cascade, but allow renewal and iteration if the final data-driven prediction models don’t perform well.

\section{Computational Process}\label{Sec:4}

\subsection{Data Preprocessing}

Icing and no-icing time periods are provided so that “normal” and “abnormal” tags are labeled for WTs 15 and 21. Table~\ref{table:distribution} listed the number of observations for each category. “invalid” means that the observations don’t belong to “normal” or “abnormal”, which could be discarded. For WT 15, there 23892 abnormal samples, 350255 normal samples, 19739 invalid samples. Similar pattern is shown for WT 21. It is obvious to see that the training data set is extremely unbalanced. Normal samples occupied most part of the data set. Under-sampling technique is used for preprocessing and make the training data set evenly distributed across different categories~\cite{lopez2013insight,tan2016wind}. As the invalid samples don’t contain useful information for icing prediction and only accounts for 0.55\% of the total samples, they are deleted.

\begin{table}[t]
\tiny
		\caption{Distribution of labeled datasets}
		\label{table:distribution}
		\centering
		\begin{tabular} {  cccc } \toprule
				\textbf{Dataset}	 &  \textbf{abnormal}	 &  \textbf{normal}	 &  \textbf{invalid} \\ \midrule
				WT 15	 &  23892	 &  350255	 &  19739   \\
				WT 21	 &  10638	 &  168930	 &  10926   \\				
				\bottomrule
		\end{tabular}
\end{table}

Another important task of data preprocessing is denoising. In industrial application, data collected by various sensors are usually noisy due to high-frequency sampling and sensor malfunctions. Wind turbine SCADA data used in this paper is collected every 7 seconds. But not all sensors, such as wind speed sensor, have such high precision. Therefore, moving average can smooth the original time series. Different steps (5, 10, 15, 25) of moving average are tried to denoise the training data. This paper finally selected 10-step moving average based on try and error. Besides moving average denoising, other sophisticated denoising methods, such as wavelet, could be used as well. But in this paper, it is not the focus to discuss what is the best way of denoising.

\subsection{Feature Engineering}

\begin{figure}
\centering
\begin{minipage}[t]{1\linewidth}
\centering
\includegraphics[width=\linewidth]{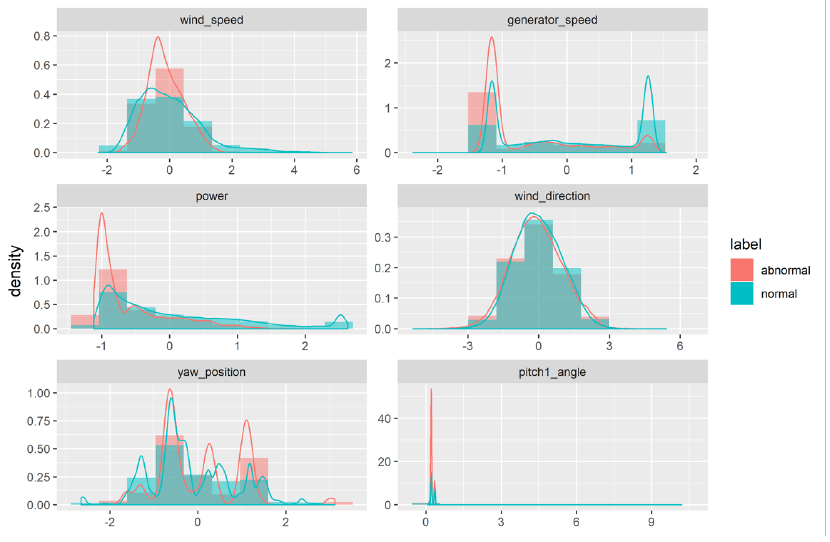}
\begin{minipage}[t]{1\linewidth}
\centering
\includegraphics[width=\linewidth]{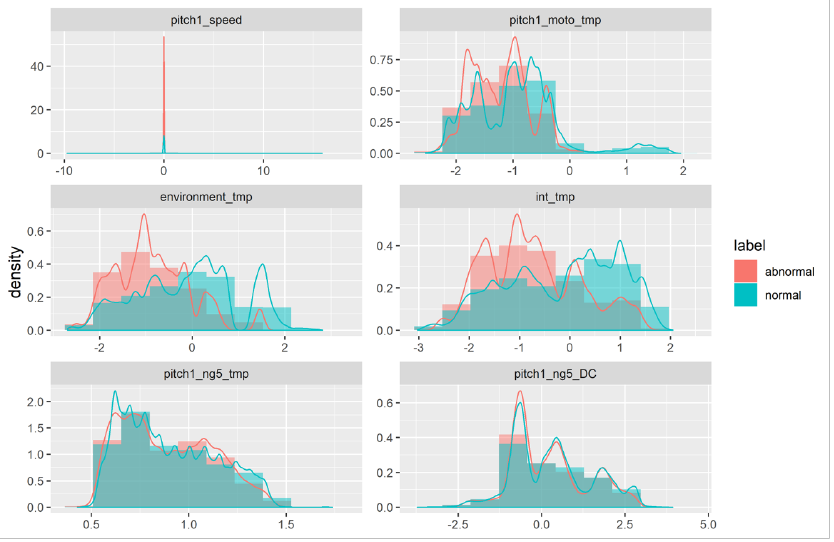}
\end{minipage}
\end{minipage}
\\ 
\caption{Density distribution of raw features for normal and abnormal samples}
\label{fig:density_distribution}
\end{figure}

Feature engineering is essential to applied machine learning problems. A good feature is worth a thousand words. By observing density distributions of original features between normal and abnormal samples, there are some features, such as wind speed, generator speed, power and pitch motor temperature, show observable differences between normal and abnormal data sets (see Fig~\ref{fig:density_distribution}). Some features are not so distinguishing in terms of icing and no-icing, such as yaw speed and acceleration, which may not have effective classification power. The goal of feature engineering by physical principle is to generate new explainable and distinguishing features

Mainstream wind turbine has three blades which should be theoretically controlled simultaneously. In the original SCADA data, each blade is equipped with sensors to monitor pitch angle, pitch speed, motor temperature and motor electrical current and so on. These monitored values are basically same across three blades. So, the average values of these features may better reflect the overall blades status than using original features independently. New features such as average pitch angle, average pitch speed, average pitch motor temperature, and so on, are generated. Temperature difference between inside and outside is generated because it is a good indicator of icing. These new features are listed in Table~\ref{table:statistical_transformation}.

\begin{table*}[htb]
\tiny
	\caption{Feature generated based on simple statistical transformation and domain knowledge}
	\label{table:statistical_transformation}
	\centering
	\setlength{\tabcolsep}{16pt}
	\scalebox{1}{
	\begin{tabular}{p{1.0cm}p{3.0cm}p{7.0cm}}
		\toprule
				\textbf{New features} &  \textbf{Description} &  \textbf{Formula}  \\ \midrule
				pitch\_angle\_Ave   &  	average pitch angle	 &  	pitch\_angle\_Ave = $\frac{(\text{pitch1\_angle} + \text{pitch2\_angle} +\text{pitch3\_angle})}{3}$     \\
				pitch\_speed\_Ave   &  	average pitch speed	 &  	pitch\_speed\_Ave =  $\frac{(\text{pitch1\_speed} +\text{pitch2\_speed} +\text{pitch3\_speed})}{3}$     \\
				pitch\_moto\_tmpAve  & average pitch motor temperature &  pitch\_moto\_tmpAve = $ \frac{(\text{pitch1\_moto\_tmp} +\text{pitch2\_moto\_tmp} +\text{pitch3\_moto\_tmp})}{3}$    \\
				pitch\_ng5\_tmpAve	 &  	average pitch Ng5 temperature	 &  	pitch\_ng5\_tmpAve =  $\frac{(\text{pitch1\_ng5\_tmp} +\text{pitch2\_ng5\_tmp} +\text{pitch3\_ng5\_tmp})}{3}$    \\
				pitch\_ng5\_DCAve	 &  	average pitch Ng5 DC	 	&  		pitch\_ng5\_DCAve =  $ \frac{(\text{pitch1\_ng5\_DC} +\text{pitch2\_ng5\_DC} +\text{pitch3\_ng5\_DC})}{3}$    \\
				tmp\_diff  &  	difference between inside and outside temperature		 &  		tmp\_diff = $\text{int\_tmp} - \text{environment\_tmp}$	   \\		
		\bottomrule
	\end{tabular}
	}
\end{table*}

Wind turbine blade icing is a typical physical phenomenon. Icing will change the shape of blades and thus undermine the aerodynamic properties~\cite{chou2013failure,jimenez2019linear}. Icing is a slow physical process where the energy is accumulated over time. Wind speed, outside temperature, humidity, wind turbine height, blade shape, rated power, rotational speed and so on, all could influence when and where the icing starts. Based on physical principle, original features are transformed into new ones, such as torque, power coefficient, thrust coefficient and tip-speed ratio. Theoretically, these new features should show good distinguishing capability. The calculation methods of these new features are show in Table~\ref{table:physical_transformation}. Because the original data is desensitization, it is necessary to add 5 to avoid 0 in the denominator. Note that the formula used here is not exactly the same as the one specified in physical textbook because some required coefficients or variables are not available or can’t be directly measured.

\begin{table}[t]
\tiny
		\caption{New feature generated based on physical principle}
		\label{table:physical_transformation}
		\centering
		\setlength{\tabcolsep}{16pt}		
		\begin{tabular} {  ccc } \toprule
				\textbf{New features} &  \textbf{Description} &  \textbf{Formula}  \\ \midrule
				torque	 & 	 torque	 & 	torque=$\frac{(\text{power}+5)}{(\text{generator\_speed} +5)}$ \\
				Cp	 & 	power factor	 & 	Cp= $\frac{(\text{power} +5)}{((\text{wind\_speed} +5)3)}$ \\
				Ct 	 & 	thrust coefficient	 & 	Ct = $\frac{\text{torque} }{\text{(wind\_speed} +5)2}$ \\
				$\lambda$	 & 	tip-speed ratio	 & 	$\lambda= \frac{(\text{generator\_speed}+5)}{(\text{wind\_speed}+5)}$ \\
				\bottomrule
		\end{tabular}
\end{table}

\begin{figure}[t]
\begin{minipage}[t]{1\linewidth}
\centering
\includegraphics[width=\linewidth]{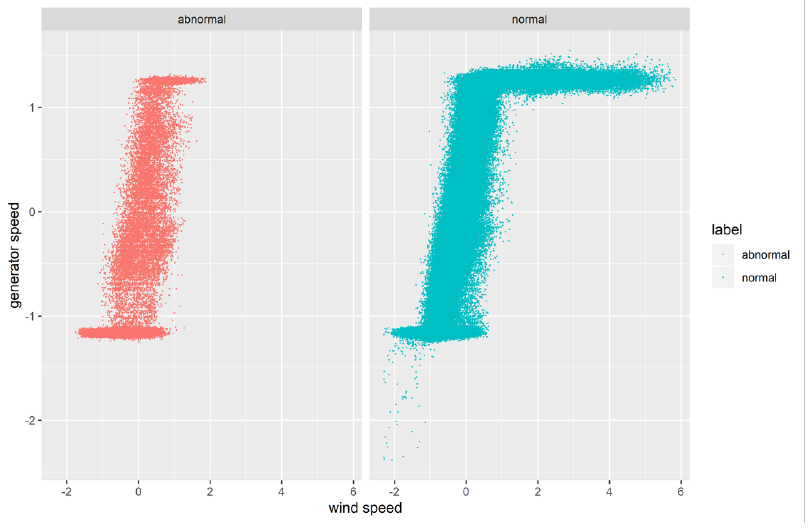}
\end{minipage}
\caption{Scatter plot of wind speed and generator speed under normal and abnormal conditions}
\label{fig:wind_speed_and_gs}
\end{figure}

\begin{figure}[t]
\begin{minipage}[t]{1\linewidth}
\centering
\includegraphics[width=\linewidth]{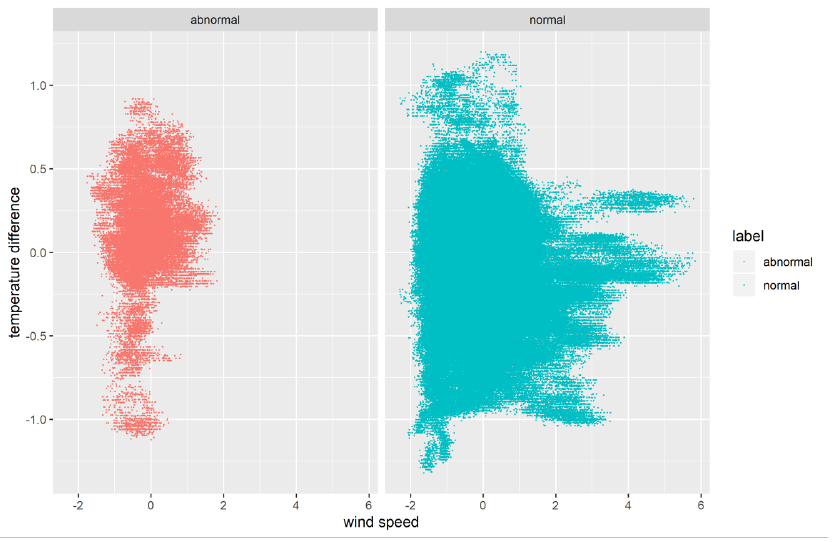}
\end{minipage}
\caption{Scatter plot of wind speed and temperature difference of normal and abnormal samples}
\label{fig:wind_speed_and_tiff}
\end{figure}

Fig~\ref{fig:wind_speed_and_gs} and Fig~\ref{fig:wind_speed_and_tiff} show scatter plots of wind speed and other features, different patterns are obvious between normal and abnormal samples. Thus, the feature transformation based on physical principle is effective.

\begin{figure}
\centering
\begin{minipage}[t]{1\linewidth}
\centering
\includegraphics[width=\linewidth]{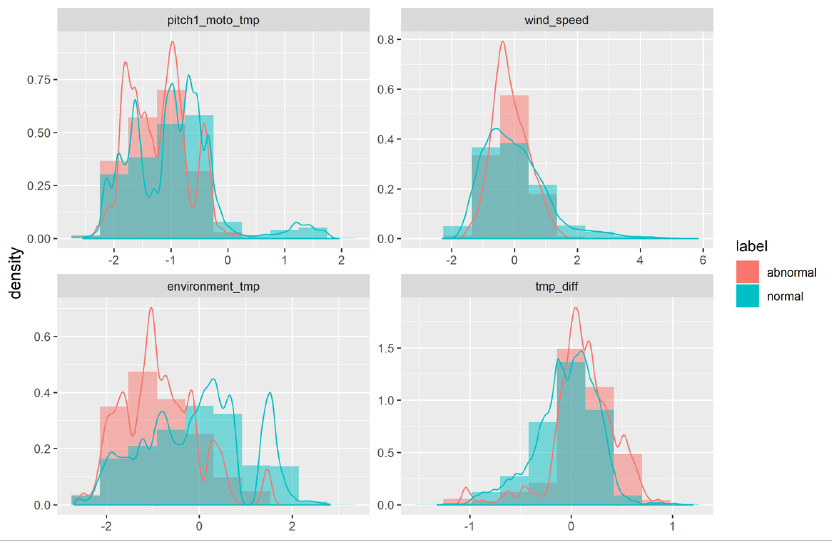}
\begin{minipage}[t]{1\linewidth}
\centering
\includegraphics[width=\linewidth]{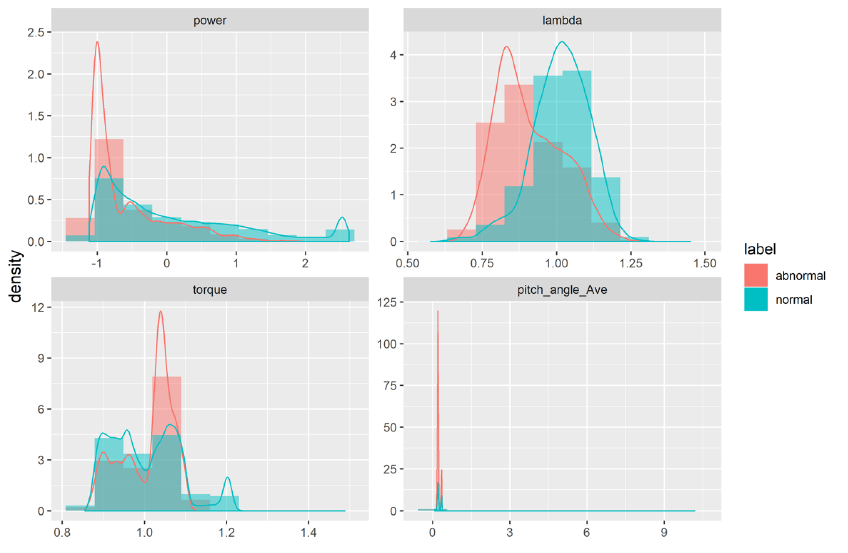}
\end{minipage}
\end{minipage}
\\ 
\caption{Density distribution of selected features under normal and abnormal conditions}
\label{fig:selected_features}
\end{figure}

The new feature and original ones are further selected based on classical feature selection algorithm to reduce prediction model’s complexity.  Final selected features are show in Table~\ref{table:final_features}. The density distributions of these features under normal and abnormal conditions are shown in Fig~\ref{fig:selected_features}.

\begin{table}[t]
\tiny
		\caption{Final selected features to build prediction model of blade icing}
		\label{table:final_features}
		\centering
		\begin{tabular} { ccc } \toprule
				\textbf{Variable} &  \textbf{Feature} &  \textbf{Description}  \\ \midrule
				$x_1$       &  	pitch1\_moto\_tmp	      &  	pitch motor 1 temperature \\
				$x_2$      &  	pitch2\_moto\_tmp	      &  	pitch motor 2 temperature \\
				$x_3$      &  	pitch3\_moto\_tmp	      &  	pitch motor 3 temperature \\
				$x_4$      &  	wind\_speed      &  		wind speed \\
				$x_5$      &  	environment\_tmp      &  		outside temperature \\
				$x_6$      &  	tmp\_diff	      &  	temperature difference between inside and outside \\
				$x_7$      &  	power      &  		power \\
				$x_8$      &  	$\lambda$      &  		tip-speed ratio \\
				$x_9$      &  	torque      &  		torque \\
				$x_{10}$      &  	pitch\_angle\_Ave	&	average  pitch angle \\
				$y$      &  	icing label	      &  	normal or abnormal \\
				\bottomrule
		\end{tabular}
\end{table}

\subsection{Strong Rule Filtering}

Based on the preliminary data exploration, blade icing mainly occurs at low wind speed, low power, low temperature and small blade angle, which can be observed in Fig~\ref{fig:wind_speed_and_gs} and Fig~\ref{fig:wind_speed_and_tiff}. Based on domain knowledge and physical principle, 5 rules could be derived and tell us under what conditions icing is going to happen. These 5 rules could effectively reduce the learning time by excluding noisy data and samples that are obviously not icing. So, the machine learning algorithm could focus on learning from data satisfying these rules.
\begin{itemize}
\item Rule 1: $x_4<2$; (IF wind speed is smaller than 2 THEN icing is possible);
\item Rule 2: $0.2 \leq x_{10} \leq 0.4$; (IF average pitch angle is between 0.2 and 0.4 THEN icing is possible);
\item Rule 3: $x_{4}<2 \  \& \  0.2 \leq x_{10} \leq 0.4$ ; (IF wind speed is smaller than 2, and average pitch angle is between 0.2 and 0.4 THEN icing is possible);
\item Rule 4:  $x_{4}<2 \  \& \  x_{5}<1.5 \  \& \  0.15<x_{10}<0.36$; (IF wind speed is smaller than 2, and average pitch angle is between 0.15 and 0.36 THEN icing is possible);
\item Rule 5: $x_{4}<2 \  \& \  x_{5}<1.5 \  \& \  0.15<x_{10}<0.36 \  \& \  x_{7}<2$; (IF wind speed smaller than 2, and outside temperature is less than 1.5, and average pitch angle is between 0.15 and 0.36, and power is less than 2 THEN icing is possible).
\end{itemize}

After several groups of try and error experiments, Rule 5 is chosen as the final strong rule for filtering.

\begin{figure}[t]
\begin{minipage}[t]{1\linewidth}
\centering
\includegraphics[width=\linewidth]{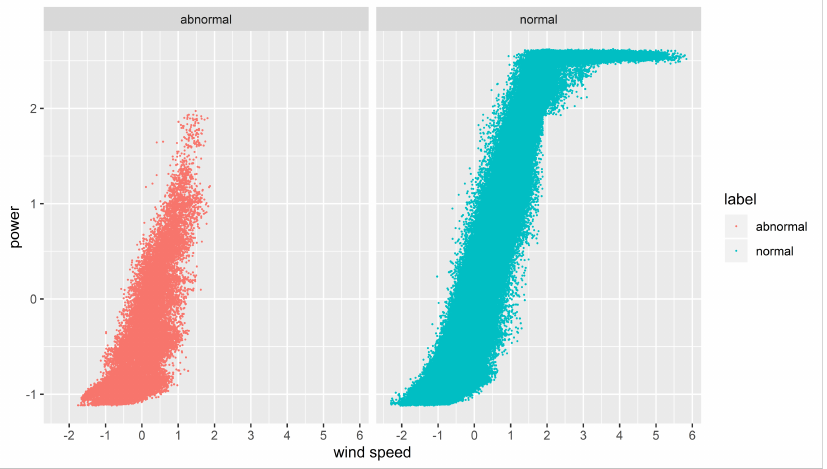}
\end{minipage}
\caption{Strong rule data visualization}
\label{fig:strong_rule}
\end{figure}

Observations outside the strong rule are normal (no-icing) data. The strong rule visualization results of the training data are shown in Fig~\ref{fig:strong_rule}. It can be observed that the outside environment temperature and wind speed are in a certain range when blade is no-icing, see bottom part of Fig~\ref{fig:strong_rule}. Fig~\ref{fig:strong_rule}'s middle part shows the power curves, i.e. scatter plot of wind speed and power, exhibit obvious patterns between icing and no-icing. When wind speed is high enough and power is large enough, the chance of blades getting frozen is little. Upper part of Fig~\ref{fig:strong_rule} (scatter plot of wind speed and average pitch angle) tells us that icing usually happened when blade pitch angle is controlled at a small value. The underlying reason is that only when wind speed is large enough, control system is activated to pitch blades to cast off extra wind energy.

\subsection{Data Segmentation}

As natural wind, the fuel, is not controllable, wind turbine is a complex energy conversion system and operated under different states according to predefined requirements. If wind speed is too high, larger than the cut-out speed, wind turbine will be shut down to keep safety. Cut-in wind speed is selected as the segmentation point based on physical principle. Below and above the cut-in wind speed, the operational logic of wind turbine is theoretically different. As the training data was desensitization, the exact cut-in wind speed is not clear. Segmentation points are mostly negative and 0, -0.25, -0.5, -0.5, -0.75, -1 are tried to segment the data into two subsets, one is higher than cut-in wind speed and the other is lower than cut-in wind speed. After several rounds of experiments the optimal segmentation point is determined at -0.25.

\section{Computational and Competition Results}\label{Sec:5}

Table~\ref{table:distribution} shows the training sample distribution before balancing treatment. Using unbalanced training data set to construct prediction model will result in poor performance. Over-sampling, under-sampling and hybrid sampling are tried to equalize the training data. Computational experiments show that the optimal equalization method is under-sampling, i.e. to randomly sample data from normal data points and put them together with icing data points to form a new training data set.

Three machine-learning classification algorithms: K Nearest Neighbors (KNN), Classification And Regression Tree (CART), and Deep Neural Network (DNN) are used for modeling. TP, TN, FN and FP (see Table~\ref{table:TP, TN, FN and FP}) are calculated respectively based on testing data set and cross-validation. The scores of different algorithms are calculated by Equation~\ref{eqn::score}, where $N_\text{normal}$ and $N_\text{fault}$ stand for the number of no-icing and icing data samples in the testing data set

\begin{equation}\label{eqn::score}
\text {Score} =\left(1-0.5 \ast  \frac{FN}{N_{\text {normal }}}-0.5 \ast  \frac{FP}{N_{\text {fault }}}\right) \ast 100
\end{equation}

\begin{table}[t]
\tiny
		\caption{Description of TP, TN, FN and FP}
		\label{table:TP, TN, FN and FP}
		\centering
		\begin{tabular} {ccccc} \toprule
                                                     &                                     & \multicolumn{2}{c}{\textbf{Predict}}               \\  \cmidrule(l){3-4}
                                                     &                                     & \textbf{Normal}     & \textbf{Fault}        \\   \midrule
		\multicolumn{1}{c}{\multirow{2}{*}{\textbf{Actual}}} & \multicolumn{1}{c}{\textbf{Normal}} & True Positive (TP)  & False Negative (FN)   \\
		\multicolumn{1}{c}{}                                 & \multicolumn{1}{c}{\textbf{Fault}}  & False Positive (FP) & True Negative (TN)  \\
		\bottomrule
	\end{tabular}
\end{table}

\begin{figure}[t]
\begin{minipage}[t]{1\linewidth}
\centering
\includegraphics[width=\linewidth]{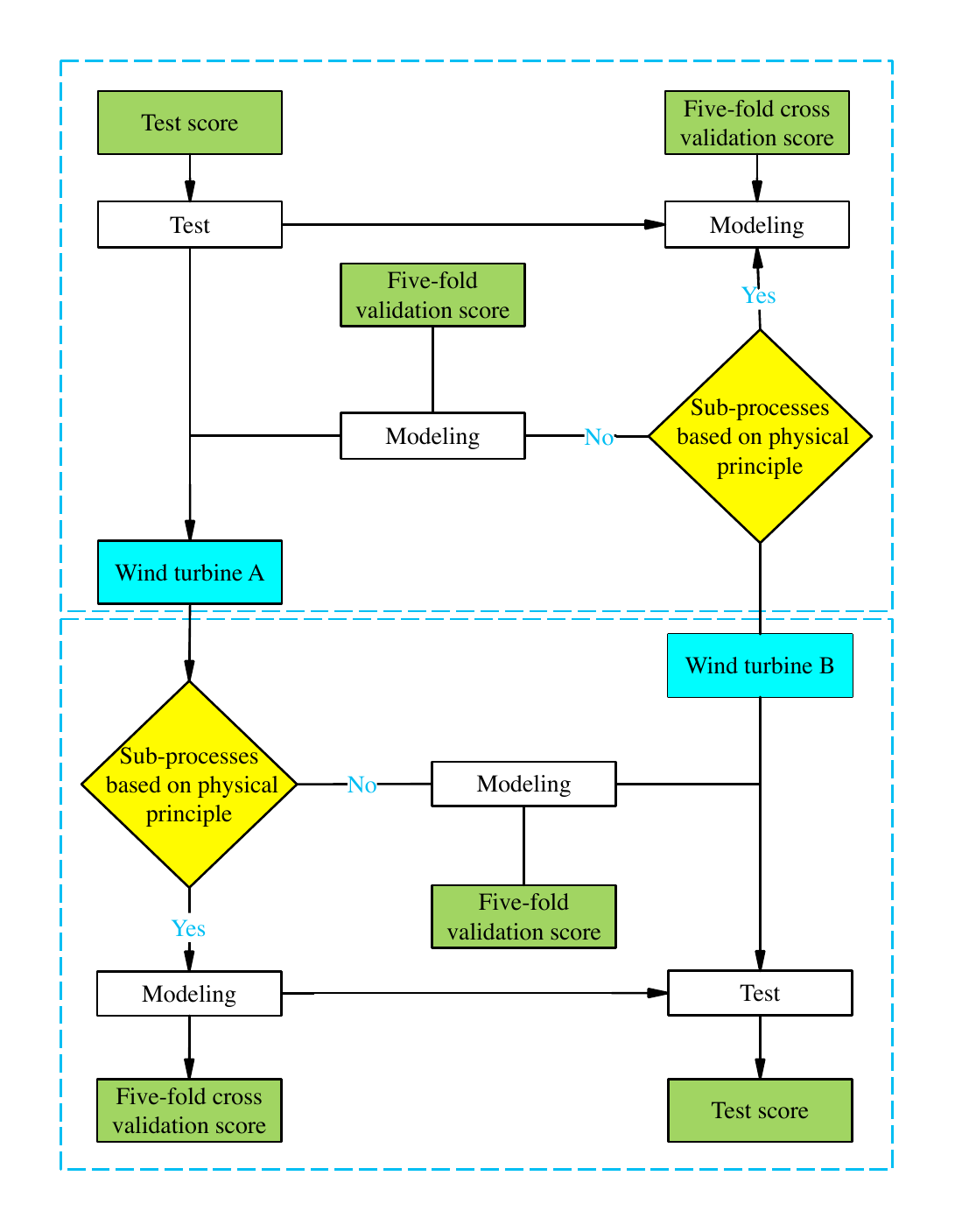}
\end{minipage}
\caption{Computational experiments for comparing re-engineered machine learning process with traditional one}
\label{fig:computational_experiments}
\end{figure}

Two sets of comparative experiments are conducted to verify the generalizability of the re-engineered machine learning process (Table~\ref{table:traditional_scores} and Table~\ref{table:re-engineered_scores}). WT 15 is for training and WT 21 is for testing, and vice versa (Fig~\ref{fig:computational_experiments}, Highlighted diamond indicates whether the training is based on the new sub-processes or not).  Learning prediction models from one turbine and test the models on another turbine is going to give the generalization performance. In order to ensure reliability of experimental results, this paper conducted ten repeated experiments, prediction models are trained on the data of ten random under-sampling, the cross-validation scores and test scores on the other wind turbine are relatively stable. The mean score and its 3standard deviation of 10 experiments with traditional machine learning process are show in Table~\ref{table:traditional_scores}. Three classical machine learning algorithms all obtained very good 5-fold cross-validation training scores. However, their test scores are very low and not acceptable for practical deployment, which means that training a prediction model from WT 15 won’t work for WT 21, and vice versa. High training scores can’t guarantee the prediction model’s generalization performance due to overfitting, multiple wind turbine running states and lack of strong features.

\begin{table}[t]
\tiny
		\caption{Scores of traditional machine learning process without physical principle integration }
		\label{table:traditional_scores}
		\centering
		\begin{tabular} {p{2.0cm}p{1.0cm}p{0.3cm}p{0.3cm}p{0.3cm}p{0.3cm}p{0.3cm}p{0.3cm}p{0.3cm}} \toprule
		                                                     &                                  & \multicolumn{2}{c}{\textbf{KNN(k=3)}}                  & \multicolumn{2}{c}{\textbf{CART}}                      & \multicolumn{2}{c}{\textbf{DNN}}                       \\   \cmidrule(l){3-8}
 		                                                    &                                  & \textbf{Mean}              & \textbf{Std.}             & \textbf{Mean}              & \textbf{Std.}             & \textbf{Mean}              & \textbf{Std.}             \\     \midrule
		                                                     & \textbf{5-fold-cross-validation} & 90.98                      & 0.44                      & 89.91                      & 0.89                      & 85.21                      & 1.07                      \\      
		\multirow{-2}{*}{\textbf{Train: WT 15  Test: WT 21}} & \textbf{Test score}              & 67.45                      & 0.43                      & 62.34                      & 7.96                      & 75.96                      & 2.57                      \\     \cmidrule(l){1-2}
		                                                     & \textbf{5-fold-cross-validation} & 89.75 & 0.70 &  95.18 &  0.62 &  86.17 &  1.43 \\   
\multirow{-2}{*}{\textbf{Train: WT 21  Test: WT 15}} & \textbf{Test score}              & 57.06                      & 0.50                      & 62.10                      & 2.36                      & 72.87                      & 2.61         \\            
		\bottomrule
	\end{tabular}
\end{table}

The mean score and its standard deviation of 10 experiments with new machine learning process are show in Table~\ref{table:re-engineered_scores}. Compared with Table~\ref{table:traditional_scores}, the new sub-processes based on physical principle can significantly improve the generalizability and accuracy of prediction models. “Data Segmentation” allows the machine learning algorithm to learn simple yet accurate models for different running states; “Strong Rule Filtering” prevents the machine learning algorithm overfitting noisy or “common-sense” samples; “Feature Engineering” generate powerful and explainable features for the machine learning algorithm.

\begin{table}[t]
\tiny
		\caption{Scores of re-engineered machine learning process with physical principle integration}
		\label{table:re-engineered_scores}
		\centering
		\begin{tabular} {p{1.8 cm}p{0.7cm}p{0.9cm}p{0.2cm}p{0.2cm}p{0.2cm}p{0.2cm}p{0.2cm}p{0.2cm}p{0.2cm}} \toprule
 				                                                  &                                      &                                  & \multicolumn{2}{c}{\textbf{KNN(k=3)}} & \multicolumn{2}{c}{\textbf{CART}} & \multicolumn{2}{c}{\textbf{DNN}} \\   \cmidrule(l){4-9}
 				                                                  &                                      &                                  & \textbf{Mean}     & \textbf{Std.}     & \textbf{Mean}   & \textbf{Std.}   & \textbf{Mean}   & \textbf{Std.}  \\   \midrule
				\multirow{4}{*}{\textbf{Train: WT 15 Test: WT 21}} & \multirow{2}{*}{\textbf{Low speed}}  & \textbf{5-fold cross validation} & 93.40             & 5.57              & 87.64           & 1.61            & 84.81           & 2.98           \\
				                                                   &                                      & \textbf{Test score}              & 77.72             & 1.39              & 83.26           & 1.68            & 76.84           & 6.62           \\  \cmidrule(l){2-3}
				                                                   & \multirow{2}{*}{\textbf{High speed}} & \textbf{5-fold cross validation} & 96.92             & 0.39              & 89.76           & 0.82            & 85.97           & 0.88           \\
				                                                   &                                      & \textbf{Test score}              & 84.21             & 0.70              & 73.19           & 2.85            & 78.39           & 4.66           \\   \cmidrule(l){1-3}
				\multirow{4}{*}{\textbf{Train: WT 21 Test: WT 15}} & \multirow{2}{*}{\textbf{Low speed}}  & \textbf{5-fold cross validation} & 96.81             & 0.42              & 87.84           & 0.76            & 86.30           & 1.46           \\
				                                                   &                                      & \textbf{Test score}              & 96.70             & 0.23              & 85.91           & 0.83            & 84.47           & 1.77           \\   \cmidrule(l){2-3}
				                                                   & \multirow{2}{*}{\textbf{High speed}} & \textbf{5-fold cross validation} & 97.03             & 0.33              & 89.47           & 0.45            & 85.95           & 0.94           \\
 				                                                  &                                      & \textbf{Test score}              & 96.78             & 0.24              & 87.93           & 0.83            & 84.72           & 2.15          \\
		\bottomrule
	\end{tabular}
\end{table}

During the competition, to ensure the generalizability of final prediction model, different random seeds are used to conduct repeated experiments, and selected the most stable model with highest prediction score, then it is deployed to predict the final test data set provided by the competition organizer. Through previous training and testing, the KNN model has highest prediction ability and generalize very well. Therefore, it is chosen as the prediction model for the preliminary testing and final competition. In the final blind test, our submitted results ranked the second place. The scores of the preliminary and final tests are shown in Table~\ref{table:Preliminary_final_scores}. The prediction model used by the  champion team is based on CNN-LSTM (Convolutional Neural Network and Long Short-Term Memory Neural Network), a deep learning algorithm, their score is 82.54, which is slightly better than our KNN’s score 82.01. However, during the final presentation in Beijing, our method is well recognized by competition judges, domain experts and industrial practitioners in terms of simplicity, computing efficiency and generalizability. As a result, our presentation score is the number one. Total runtime is about 555.6 seconds, including raw data preprocessing 160.34 seconds, feature engineering 384.72 seconds, model training 9.06 seconds, new data set testing 2.24 seconds. Running environment is 64bit-Win10, CPU i7-2600 3.4GHz, RAM 8G. The R code (448 lines) is complied with R version 64bit-3.3.2. Other contestants, i.e., the first-place team with deep learning algorithms spent more than 6 hours on preprocessing and training with similar computers.

\begin{table}[t]
\tiny
		\caption{Preliminary and final scores with rankings}
		\label{table:Preliminary_final_scores}
		\centering
		\begin{tabular} {ccccc} \toprule
				\textbf{Blind Test} &  \textbf{Dataset} &  \textbf{Score}  &  \textbf{Algorithm} &  \textbf{Ranking}  \\ \midrule
				Preliminary    &  WT 8 	&  	89.23	 	&  KNN (K=3) 	&  2   \\
				Final    &  WT 10, WT 14		 &  	82.01		 &  	KNN (K=3)	 &  	2     \\
		\bottomrule
	\end{tabular}
\end{table}

\section{conclusion}\label{Sec:6}
Like manufacturing an industrial equipment, this paper treats machine learning as a production process and final product is prediction model. In order to improve prediction model’s generalizability, this paper re-engineered the traditional machine learning process by integrating physical principle. Computational experiments and blind-test competition based on real-world industrial data sets prove the effectiveness of our methodology. This paper also shows that simple machine learning algorithms can compete with deep learning algorithms with the new process in industrial setting, which may shed light on an even bigger research question, is the learning process important or the learning algorithm important? Recent fast development of deep learning algorithms and their successful applications in image or speech recognition are in deep contrast with limited predictive capabilities in industrial fields. It is time to rethink machine learning in a systematic way, where high-quality data, algorithms, processes and later maintenance are all essential to successful deployment of industrial data-driven prediction models. However, this research is limited in covering depth and width of machine learning process engineering. Other sub-processes and life-cycle of prediction model management are not discussed, and could be our future research directions. Data-driven predictive models for different industries, different equipments and systems may need different machine learning processes, which is not fully researched in this paper and worthy of further investigation.

\IEEEpeerreviewmaketitle

\bibliographystyle{IEEEtran}
\bibliography{biblio}

\end{document}